\documentclass[conference]{IEEEtran}
\IEEEoverridecommandlockouts

\usepackage{tikz}
\usepackage{textcomp}
\usepackage{lipsum}

\newcommand\copyrighttext{%
  \footnotesize \textcopyright 2023 IEEE. Personal use of this material is permitted. However, permission to reprint/republish this material for advertising or promotional purposes or for creating new collective works for resale or redistribution to servers or lists, or to reuse any copyrighted component of this work in other works must be obtained from the IEEE.}
\newcommand\copyrightnotice{%
\begin{tikzpicture}[remember picture,overlay]
\node[anchor=south,yshift=10pt] at (current page.south) {\fbox{\parbox{\dimexpr\textwidth-\fboxsep-\fboxrule\relax}{\copyrighttext}}};
\end{tikzpicture}%
}

\usepackage{cite}
\usepackage[hidelinks]{hyperref}
\usepackage[nottoc]{tocbibind}
\usepackage{amsmath,amssymb,amsfonts}
\usepackage{algorithmic}
\usepackage{graphicx}
\usepackage{textcomp}
\usepackage{xcolor}
\def\BibTeX{{\rm B\kern-.05em{\sc i\kern-.025em b}\kern-.08em
    T\kern-.1667em\lower.7ex\hbox{E}\kern-.125emX}}

\usepackage{algorithmic,algorithm}
\usepackage{mathtools}
\usepackage{enumitem}
\usepackage{arydshln}
\usepackage{caption}
\DeclarePairedDelimiter\ceil{\lceil}{\rceil}

\begin{document}

\title{Snapshot Spectral Clustering --\\ a
costless approach to deep clustering ensembles generation
}

\author{
\IEEEauthorblockN{Adam Piróg\textsuperscript{*}} \\
\IEEEauthorblockA{\textit{4Semantics, Warsaw, Poland \vspace{2mm}} \\ 
\textit{Department of Artificial Intelligence -- } \\
\textit{Wrocław University of Science and Technology,} \\
\textit{Wrocław, Poland}
}

\and
\IEEEauthorblockN{Halina Kwaśnicka} \\
\IEEEauthorblockA{\textit{ \vspace{2mm}} \\
\textit{Department of Artificial Intelligence -- } \\
\textit{Wrocław University of Science and Technology,}}
\textit{Wrocław, Poland}
}

\maketitle
\copyrightnotice

\let\oldfoot=\thefootnote

\begingroup\renewcommand\thefootnote{*}
\footnotetext{Corresponding author. E-mail: a.pirog@4semantics.pl}

\let\thefootnote=\oldfoot
\begin{abstract}
    Despite tremendous advancements in Artificial Intelligence, learning from large sets of data in an unsupervised manner remains a significant challenge. Classical clustering algorithms often fail to discover complex dependencies in large datasets, especially considering sparse, high-dimensional spaces. However, deep learning techniques proved to be successful when dealing with large quantities of data, efficiently reducing their dimensionality without losing track of underlying information. Several interesting advancements have already been made to combine deep learning and clustering. Still, the idea of enhancing the clustering results by combining multiple views of the data generated by deep neural networks appears to be insufficiently explored yet. This paper aims to investigate this direction and bridge the gap between deep neural networks, clustering techniques and ensemble learning methods. To achieve this goal, we propose a novel deep clustering ensemble method – Snapshot Spectral Clustering, designed to maximize the gain from combining multiple data views while minimizing the computational costs of creating the ensemble. Comparative analysis and experiments described in this paper prove the proposed concept, while the conducted hyperparameter study provides a valuable intuition to follow when selecting proper values.
\end{abstract}
\section{Introduction}
Although Machine Learning is a relatively young field of science, it has made tremendous progress in recent years, surpassing people in many areas. AI models taught in a supervised manner played a key role in this advancement, but they still suffer from certain drawbacks. Creating labelled datasets remains time-consuming and expensive, especially considering tasks requiring highly qualified professionals (e.g. labelling medical images). This is why adapting unsupervised techniques, such as clustering, to large-scale problems is important.
\par 
While classical clustering algorithms often fail to discover complex dependencies in large datasets, deep learning techniques proved to be successful when dealing with large quantities of data. They can efficiently reduce data dimensionality without losing track of underlying information. Combining clustering with deep learning seems a natural direction. Still, the idea of enhancing the clustering results by combining multiple views of the data generated by neural networks appears to be insufficiently explored yet. 
\par
In this paper, we propose a \textbf{novel deep clustering ensemble method} – Snapshot Spectral Clustering (SSC), designed to maximize the gain from combining multiple data views while minimizing the computational costs of creating the ensemble. The SSC algorithm, with its innovative solutions reducing time and memory requirements of creating the ensemble and enhancing the diversity of its members is the main contribution.
\par
Our paper presents \textbf{proof of concept of the proposed method}. Instead of comparing Snapshot Spectral Clustering to state-of-the-art models, which require using a backbone network with a comparable number of trainable parameters, we conducted experiments showing the proposed method's properties and potential power. Such experiments show the strengths and weaknesses of this approach and suggest the direction of further research without involving huge computing power. 
\par 
The remainder of this paper is organized as follows: in Section~\ref{sec:method}, we present the SSC method; in Section~\ref{sec:experiments} we describe the experimental and protocol; in Section~\ref{sec:results} we discuss the results; in Section~\ref{sec:conclusion} we summarize our research and outline directions for future work.

\subsection*{Related Work}
Several interesting advancements have been made to combine clustering with deep learning techniques by taking different approaches. DEC~\cite{dec} has been one of the first deep clustering algorithms -- it uses features learned from an autoencoder to fine-tune the cluster assignments. Other works~\cite{end1, end2, end3} have shown that it is possible to achieve high results when training an architecture with an end-to-end clustering base objective. Another interesting approach, taken by many papers~\cite{deep-cluster, spice, self-label}, is to train deep neural networks with pseudolabels generated by a clustering algorithm. Recently, the field has also seen a tremendous advancement~\cite{contrastive1, contrastive2} by incorporating contrastive representation learning to enhance clustering.  
\par 
While several ensemble clustering algorithms~\cite{consensus1, consensus2} are well known and tested, the connection between deep neural networks, clustering techniques and ensemble learning methods seems to be insufficiently explored yet. A paper proposed by Affeldt~et~al.~\cite{sc-edae} investigates this exact idea but still leaves multiple advancements to be made. Authors of SC-EDAE train multiple autoencoders with different hyperparameters, merge the embeddings into a single affinity matrix to finally partition it with a Spectral Clustering algorithm. The idea is effective, yet, suffers from certain drawbacks our algorithm aims to address (primarily high costs of ensemble generation). 
\section{Proposed method -- Snapshot Spectral Clustering (SSC)}
\label{sec:method}
The general objective of the algorithm can be formulated as follows:
%
given data matrix $X \in R^{n \times d}$, where \textit{n} is the size, and \textit{d} is the dimensionality of the dataset, the goal is to produce \textit{k} data partitions, which reflect dependencies in underlying data.
\par
In the first step of the SSC algorithm, we obtain \textit{m} embeddings of the data. Subsequently, we construct the affinity matrix $\overline{Z}$, which preserves information from each of the \textit{m} encodings. Finally, the clustering algorithm is applied on matrix  $\overline{Z}$ to produce the final result of \textit{k} partitions.
\par
\noindent \noindent
\textbf{The main shortcomings arising from the analysis of relevant literature that SSC method is trying to address are}:
\begin{enumerate}
    \item to reduce high training costs of deep clustering ensembles, such as SC-EDAE~\cite{sc-edae},
    \item to reduce high memory requirements of Spectral Clustering while maintaining its descriptive abilities,
    \item to utilize the properties of LSC~\cite{lsc}, to achieve greater diversity among base members.
\end{enumerate}
\par
What is important to note is that the specifics of the base model of the ensemble are not a part of the ensemble algorithm itself. As with every method of this type, details of the base member should remain behind a layer of abstraction -- any model can be used as long as it meets certain requirements. In the case of the SSC algorithm, the base member should be a deep learning model trained unsupervised to extract meaningful features from the data. In the following subsections we describe each step of the algorithm in details: data embedding generation, base affinity matrices and ensemble fusion process. Fully formulated SSC Algorithm ends this section.
\subsection{Data embeddings generation}
\label{sec:embedding_generation}
The process of generating multiple encodings of the data is inspired by the Snapshot Ensemble algorithm, proposed by Huang~et~al.~\cite{snapshot_ensemble}.
\par
The backbone neural network is trained using a Cosine Learning Scheduler, which allows the creation of the ensemble of encoders by reaching multiple local minima of the loss function. The learning rate at each iteration $\alpha (t)$ is given by the following formula:
$$
\alpha(t) = \frac{\alpha_0}{2}(cos(\frac{\pi mod(t - 1, \ceil{T/M})}{\ceil{T/M}}) + 1)
$$
\noindent \noindent
hyperparameters of the method being:
\begin{itemize}
   \item $\alpha_0$ the initial/maximal learning rate
    \item \textit{T} the total number of training iterations
    \item \textit{M} the number of cycles/snapshots
\end{itemize}
\par
\noindent \noindent
While training the network in this manner, a ''snapshot'' of encoders weights is saved after $\frac{T}{M}$ epochs, serving as the member of the ensemble. Then the encoding $Y \in R^{n \times d'}$, where $d'$ denotes the size of the encoding layer, is extracted from each member and is further used to create an affinity matrix of the data, as described in the next section. 
\par
As proven by multiple papers~\cite{snapshot_suplement, local_minima}, the local minima of the model are not far from the global minimum in the case of networks performance. Models created using this technique will not be subpar to a theoretical model trained with the same budget in a traditional way~\cite{snapshot_ensemble, snapshot_suplement}, at the same achieving significant diversity between members. 
\subsection{Obtaining base affinity matrices}
\label{sec:base_affinity_matrix}
The process of obtaining a base affinity matrix from each encoding generated during the snapshot training process is based on LSC algorithm~\cite{lsc}.
\par
To avoid creating a full affinity matrix $Z \in R^{n \times n}$, needed for classical Spectral Clustering, we define \textit{p} landmarks in the data space -- representative points chosen to approximate the neighbourhood structure (\textit{p} is one of the hyperparameters of SSC). To obtain those landmarks efficiently, the MiniBatch KMeans algorithm is run, and after convergence, the cluster centres (centroids) are extracted to serve as landmarks. Because \textit{p} will be larger than values usually used as \textit{k}, a significant acceleration can be expected when using the MiniBatch version of KMeans.
\par
A much smaller representation matrix $\hat{Z} \in R^{n \times p}$ ($p << n$) is created to approximate full  $Z \in R^{n \times n }$ affinity matrix, using mapping between landmarks $\{u_{j} \hspace{1mm}j \in [1,p]\} $ and data points $\{x_{i} \hspace{1mm} i \in [1,n]\}$. In a further step, the sparsity of $\hat{Z}$ matrix is enforced by zeroing all values corresponding to landmarks that are not within \textit{r} nearest landmarks around given point \textit{x} -- value \textit{r} is another hyperparameter of the method, further called \textit{sparsity treshold}\footnote{Resulting matrix is very sparse (in practice around 1\% density). Thus the usage of data formats specially designed for storing sparse matrices in memory is strongly advised.}. Elements $\hat{z_{ij}}$ of similarity matrix  $\hat{Z}$ are computed as follows:
$$ 
\hat{z_{ij}} = \frac{K(x_i, u_i)}{\sum_{j' \in N_{(i)}}K(x_i, u_{j'})}
$$
\noindent\noindent
where $N(i)$ denotes the $r$  ($r < p$) nearest landmarks around point $x_i$ and $K(\cdot)$ is used to measure similarity between point $x_i$ and landmark $u_j$. The following kernel is used as a similarity $K(\cdot)$:
$$ 
    s(x_i, u_i) = exp(\frac{-dist(x_i, u_i)^2}{2\sigma^2})
$$
\noindent \noindent
where $\sigma$ is the bandwidth parameter\footnote{Here, a rule proposed by~\cite{scott} can be used to automatically determine proper $\sigma$ value.} and \textit{dist} is a distance metric.
\subsection{Ensemble fusion process}
\label{sec:fusion_proc}
Matrices obtained by transforming generated data embeddings to landmark-base representation are further fused into single affinity matrix $\hat{Z}$ by concatenation, as proposed by Affeldt~et~al.~\cite{sc-edae}:
$$
\overline{Z} = \frac{1}{\sqrt{m}}[\hat{Z}_1 | \hat{Z}_2| ... | \hat{Z}_m]
$$
The final affinity matrix created using this method has a dimensionality of $\overline{Z} \in R^{n \times pm} \hspace{1mm} (n >> pm)$. This step further enhances members' diversity, as each base affinity matrix uses a different set of landmarks to describe the data. From this matrix, we extract \textit{k} left singular vectors, as they share properties with eigenvectors of underlying graph Laplacian~\cite{sc-edae}, to create a matrix $U \in R^{n \times k}$. 
\par
Lastly, the KMeans clustering algorithm is used on matrix \textit{U} to obtain the final result of \textit{k} data partitions. 
\subsection{SSC Algorithm}
\label{sec:ssc_alg_desc}
The proposed method (presented as Algorithm~\ref{alg1}) addresses all the goals formulated at the beginning of this section:
\par
Utilizing Snapshot Ensemble methods to create base members of the ensemble and generate diverse encodings of the data significantly reduces training costs, especially compared to the SC-EDAE method. When in SC-EDAE, each model is trained independently, SSC allows to create \textit{m} base models with the same epoch budget as creating one model, thus reducing training cost \textit{m} times. Other small improvements, such as using the MiniBatch KMeans instead of the classic version to obtain landmarks, allowes for further algorithm acceleration.

\begin{algorithm}[ht]
\small
\caption{Snapshot Spectral Clustering}
\label{alg1}
\begin{enumerate}[leftmargin=*]
    \item Generate \textit{m} embeddings $Y \in R^{n \times d'}$ of the original data $X \in R^{n \times d}$ using Snapshot Ensemble method, as described in~\ref{sec:embedding_generation}.
    \item For each embedding, create an affinity matrix $\hat{Z} \in R^{n \times p}$ , using landmark-based representation, as described in~\ref{sec:base_affinity_matrix}. 
    \item Merge all matrices into single affinity matrix  $\overline{Z} \in R^{n \times mp}$,\\as  as described in~\ref{sec:fusion_proc}.
    \item Extract \textit{k} left singular vectors from $\overline{Z}$ creating $U \in R^{n \times k}$.
    \item Run KMeans algorithm on \textit{U} to obtain the final result.
\end{enumerate}
\end{algorithm}
\par
Transforming base embeddings using landmark-based representation and utilizing an efficient matrix fusion method leads to an outstanding reduction of memory requirements compared to the classic Spectral Clustering algorithm. Enforcing the sparsity of affinity matrix also allows to take advantage of special mechanisms designed for storing sparse matrices in memory, reducing those requirements even further.

\par
To utilize the properties of LSC to achieve greater diversity among base members, a variant of the SSC method, called SSC-RM (presented as Algorithm~\ref{alg2}), is proposed. Using a random distance metric in each base affinity matrix allows us to take advantage of different views of the data, further enhancing the diversity among base members. Another benefit of this variant of SSC is the exclusion of one hyperparameter from the tuning process.
\begin{algorithm}[ht]
\small
\caption{Snapshot Spectral Clustering -- \small{Random Metric}}
\label{alg2}
\begin{enumerate}[leftmargin=*]
    \item Generate \textit{m} embeddings $Y \in R^{n \times d'}$ of the original data $X \in R^{n \times d}$ using Snapshot Ensemble method, see section~\ref{sec:embedding_generation}
    \item For each embedding, create an affinity matrix $\hat{Z} \in R^{n \times p}$ , using landmark-based representation, as described in~\ref{sec:base_affinity_matrix}.
    \textbf{For each affinity matrix $\hat{Z}$ use different distance metric \textit{d}.}
    \item Merge all matrices into single affinity matrix  $\overline{Z} \in R^{n \times mp}$, \\as described in~\ref{sec:fusion_proc}.
    \item Extract \textit{k} left singular vectors from $\overline{Z}$ creating $U \in R^{n \times k}$.
    \item Run KMeans algorithm on \textit{U} to obtain the final result.
\end{enumerate}
\end{algorithm}
\section{Experiments}
\label{sec:experiments}
The main goal of experiments is to deliver a proof of concept for the proposed SSC method. Further, we want to analyze the model's behaviour in different setups, show its strengths and weaknesses and provide intuition to follow when selecting hyperparameters' values. To achieve those goals, a comparison of the following models is necessary:

{
\renewcommand{\arraystretch}{1.05}
\begin{table}[ht]
\centering
\caption{List of compared models}
\begin{tabular}{ll}
\hline
\textbf{No.} & \textbf{Model}\\ \hline
1.   & KMeans \\
2.   & DAE\_KMeans \\
3.   & LSC \\
4.   & DEA\_LSC \\
5.   & SSC \\
6.   & SSC-RM   
\end{tabular}

\label{tab:compared models}
\end{table}
}

DAE\_KMeans and DAE\_LSC models work by performing clustering (with KMeans and LSC algorithms, respectively) on latent features extracted from the middle layer of the deep autoencoder. Comparison between SSC (proposed ensemble method) and DAE\_LSC (base model of the ensemble) allows us to answer the fundamental question: does the SSC ensemble outperform its base model? KMeans-based models are also included in the comparison, as it is the most widely used clustering algorithm.

Datasets used in experiments are publicly available and widely used as benchmarks for various machine-learning tasks. They consist of visual data, yet they present different levels of challenge, so the proposed method can be tested in various environments. 

MNIST~\cite{mnist} --  70,000 images of handwritten digits. All the pictures are converted into grayscale and normalized to fit into a 28$\times$28 pixel bounding box. As the pictures are small, contain only 1 channel (grayscale) and do not have any background noise -- this dataset should be a relatively easy challenge for the clustering algorithm.

CIFAR-10~\cite{cifar10} -- 60,000 32$\times$32px colour pictures in 10 different classes.  Compared to MNIST, it has bigger pictures, containing 3 channels. Above that, the pictures show a wide range of objects, varying from small animals to big vehicles, in different surroundings. Strong background, sometimes dominating the picture content, can further prove to be a key element in the cluster formation process.
\subsection{Experimental procedure}
The experimental procedure for both datasets is as follows:
\begin{enumerate}
    \item Hyperparameter study is conducted to examine models' behaviour in different setups and to provide a general intuition to follow when choosing default values.  As with the network size, differences in datasets require different domains to be used -- exact values selected for the process are coming from domains described in Table~\ref{tab:archi}. Two groups separated by the dashed line are the hyperparameters of the ensemble generation process and consensus function, respectively.
{
\renewcommand{\arraystretch}{1.1}
\begin{table}[ht]
\centering
\caption{Hyperparameters domain}
\vspace{1mm}
\begin{tabular}{ll}
\hline
\multicolumn{2}{c}{\textbf{MNIST}}                              \\ 
\textbf{Hyperparameter} & \textbf{Domain}                       \\ \hline
Cycle length            & 15, \textbf{20}, 25                   \\
Max learning rate       & 0.007, \textbf{0.01}, 0.03            \\
Encoding size           & 128, \textbf{256}, 512                \\ \hdashline
Number of landmarks     & \textbf{350}, 600, 1000               \\
Sparsity treshold       & \textbf{3}, 7, 15                     \\
Distance metric         & \textbf{euclidean}, cosine, minkowski \\ \hline
\multicolumn{2}{c}{\textbf{CIFAR-10}}                           \\ 
\textbf{Hyperparameter} & \textbf{Domain}                       \\ \hline
Cycle length            & 20, \textbf{40}, 60                   \\
Max learning rate       & 0.1, \textbf{0.2}, 0.3                \\
Encoding size           & 512, \textbf{1024}, 2048              \\ \hdashline
Number of landmarks     & 350, \textbf{600}, 1000               \\
Sparsity treshold       & 3, \textbf{7}, 15                     \\
Distance metric         & \textbf{euclidean}, cosine, minkowski
\end{tabular}
\label{tab:archi}
\end{table}
}
\noindent \noindent
Each hyperparameter is tested in a separate experiment (using all values from the respective domain). During that experiment, other hyperparameters remain fixed (defaulting to the value highlighted in bold). Each of the tests is repeated five times.

\item As the next step, a comparative study between all models, listed in Table~\ref{tab:compared models}, is conducted. Snapshot Spectral Clustering result is obtained using the best hyperparameters determined during the experiments described above and an ensemble of size 6. For a reliable, all models based on neural networks are given the same architectures and training budget. The number of epochs depends on the best value of the Snapshot cycle length: $epochs  = cycle\_length \cdot ensemble\_size$. Each of the tests is repeated five times.
\end{enumerate}
\subsection{Models architecture and training process}
\begin{figure*}[ht]
  \includegraphics[width=\textwidth,height=7.5cm]{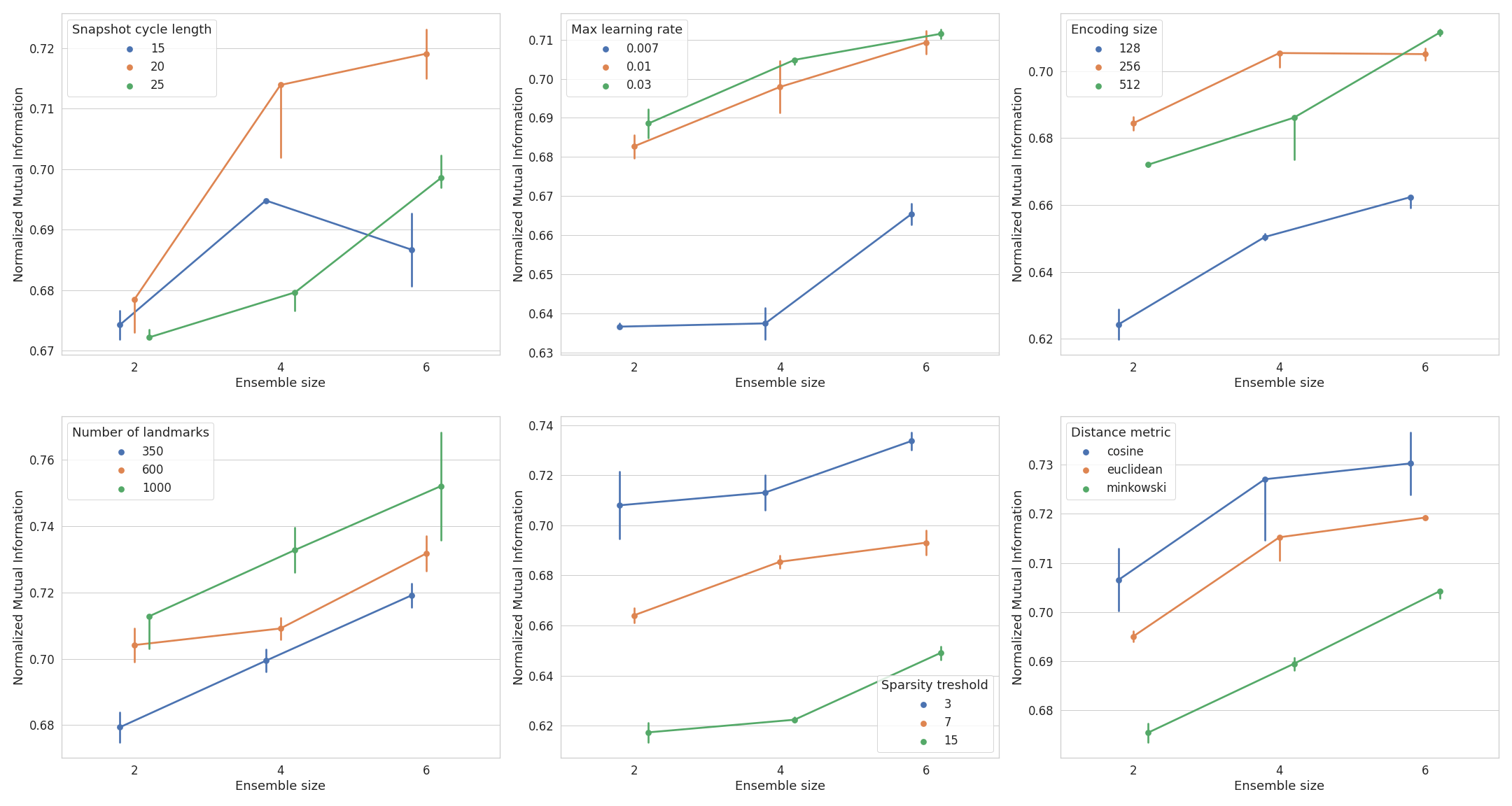}
  \vspace{-5mm}
  \caption{Hyperparameter Study}
  \label{fig:hparams}
\end{figure*}
The trained model is an undercomplete, convolutional autoencoder, with additional Gaussian noise injected in the input layer during training. In the ensemble generation process encoder is trained jointly with the decoder (built with Transposed Convolutional layers), but only encoder weights are saved for later stages of the algorithm. 
\par
For a reliable comparison, both ensembles (SSC and SSC-RM) and DAE-based methods (DAE\_KMeans, DAE\_LSC) are based on the same architecture and are trained with the same epoch budget. All of the models use Mean Squared Error as the loss function. Ensemble models are trained using Stochastic Gradient Descent (SGD) with a cosine learning scheduler. In contrast, DAE-based methods are trained using Adam optimizer~\cite{adam} with an early stopping mechanism to ensure their best possible performance.
\par
Because the datasets used in the experiment present different levels of challenge, there are separate network architectures designed for each of them. Both models follow the same general structure but differ in size and the number of layers (see Table~\ref{tab:hyper}). \textit{Conv block} represents a Convolutional layer with Batch Normalization and ReLU activation.
{
\renewcommand{\arraystretch}{0.9}
\begin{table}[ht]
\centering
\caption{Encoders Details}
\begin{tabular}{lrrr}
\hline
\multicolumn{4}{c}{\textbf{MNIST}}                                                                                                   \\ 
\textbf{Layer}                                & \textbf{Size}                                & \textbf{Kernel size} & \textbf{Stride} \\ \hline
Conv block                                    & 32                                           & 3                    & 2               \\
Conv block                                    & 32                                           & 3                    & 1               \\
Conv block                                    & 64                                           & 3                    & 3               \\
Dense                                         & 256                                          & -                    & -               \\ \hdashline
\multicolumn{2}{c}{\textbf{\begin{tabular}[c]{@{}c@{}}Trainable \\ parameters\end{tabular}}} & \multicolumn{2}{c}{420,672}            \\ \hline
\multicolumn{4}{c}{\textbf{CIFAR-10}}                                                                                                 \\ 
\textbf{Layer}                                & \textbf{Size}                                & \textbf{Kernel size} & \textbf{Stride} \\ \hline
Conv block                                    & 32                                           & 3                    & 1               \\
Conv block                                    & 32                                           & 3                    & 2               \\
Conv block                                    & 64                                           & 3                    & 1               \\
Conv block                                    & 64                                           & 3                    & 2               \\
Conv block                                    & 128                                          & 3                    & 1               \\
Conv block                                    & 128                                          & 3                    & 2               \\
Conv block                                    & 256                                          & 3                    & 1               \\
Conv block                                    & 256                                          & 3                    & 2               \\
Conv block                                    & 512                                          & 3                    & 1               \\
Dense                                         & 1024                                         & -                    & -               \\ \hdashline
\multicolumn{2}{c}{\textbf{\begin{tabular}[c]{@{}c@{}}Trainable \\ parameters\end{tabular}}} & \multicolumn{2}{c}{4,444,224}         
\end{tabular}
\label{tab:hyper}
\end{table}
}

Both the quality and sheer size of the base model can prove to be crucial for the methods' performance, especially when comparing it to other models. Yet, the number of trainable parameters is the main bottleneck defining resources required for experiments. Relatively simple architectures (especially compared to state-of-the-art models) have been chosen to limit the required computational power. Still, they should remain sufficient for delivering a proof of concept for the proposed method, which is the main goal of this study.  

\section{Results}
\label{sec:results}
Beyond quantitative evaluation, based on standard clustering performance metrics, qualitative analysis is conducted to better understand the methods' behaviour. As datasets chosen for this study present different levels of challenge, different architectures and hyperparameter domains are tested for each of them. Hyperparameter study considers two important categories: ensemble generation and consensus function hyperparameters.

\subsubsection*{Ensemble generation hyperparameters}
Results regarding encoding size, seen in Figure~\ref{fig:hparams}, suggest a general tendency towards a larger size of encoder features. In small ensembles, the encoding size of 256 seems to outperform bigger counterparts, but this trend reverses starting from ensembles of size 6. Besides providing more descriptive space, a larger encoding size allows for greater diversity among base members, which is crucial for higher performance. Considering all those factors, leaning towards a bigger encoding size seems to be a reasonable recommendation.

\par 
Hyperparameter study also reveals the best values for cosine learning scheduler -- snapshot cycle length of 20 and maximal learning rate of 0.003. Nevertheless, determining the right values for those hyperparameters is more complicated and conducted experiments do not provide a strong trend or intuition to follow. As seen in Figure~\ref{fig:hparams}, improper selection of cycle length can even deteriorate the ensemble's performance when increasing its size. Choosing a learning rate that is not big enough also results in a visible drop in model performance. What introduces even a bigger challenge is the fact that those parameters are heavily data-dependent. They strongly rely not only on size or type of data but even semantic details, which may make it difficult to work with this method on a new dataset. 

\par
Proper values for cosine learning scheduler can be determined by repeated trials (for example, using linear evaluation protocol). Still, considering the high importance of those hyperparameters and the substantial amount of resources required for such tuning, alternative approaches should also be considered\footnote{Wen, Long et al.~\cite{cosine_scheduler} propose a \textit{logLR} test method to estimate proper parameters for learning scheduler.}.

\subsection*{Consensus function hyperparameters}
\label{sec:mnist_consensus_hyperparams}

The experiment suggests that a larger number of landmarks positively impacts the model's performance, as seen in Figure~\ref{fig:hparams}. This behaviour was to be expected, as a larger matrix should code more information. Still, this hyperparameter significantly impacts the time complexity of the consensus function, so this factor should be considered in the trade-off between performance and speed. 

\par
The sparsity threshold follows the opposite trend, where the lowest value yields the best results. This can be explained by the fact that MNIST is a relatively simple dataset -- leaving the affinity matrix more dense introduces more noise than actual information. In bigger datasets, however, leaving higher sparsity values may be advised.

\par
Experiments regarding distance metrics show cosine metric achieving the best results. Euclidean distance metric still remains a close second (especially considering the confidence interval). That behaviour suggests that the Snapshot Spectral Clustering version with randomized metric could deliver strong results compared to the base version.

\subsection{MNIST}
As expected, all KMeans variants perform poorly in comparison to other methods. The algorithm cannot capture the complexity of the underlying data manifold. LSC-based models perform visibly better, with around 15 percentage points advantage. 

\begin{figure}[ht]
  \includegraphics[width=0.5\textwidth]{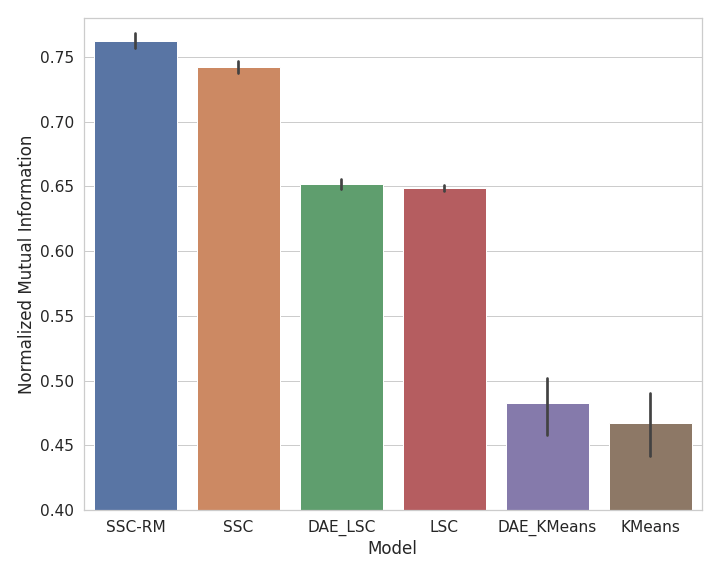}
  \vspace{-5mm}
  \caption{MNIST -- Models comparison}
  \label{fig:mnist_model_comparison}
\end{figure}

The most important comparison, in terms of this study goal, is between Snapshot Spectral Clustering and DAE\_LSC. It reveals a clear advantage of the proposed ensemble method. Above that, the randomized metric (SSC-RM) version delivers even better results than the basic version of SSC. As hyperparameters tests have shown, different distance metrics can capture valuable information about the data from different perspectives. A comparison between SSC and SSC-RM confirms that an ensemble can benefit from such diversity. 

\begin{figure}[ht]
  \includegraphics[width=0.5\textwidth]{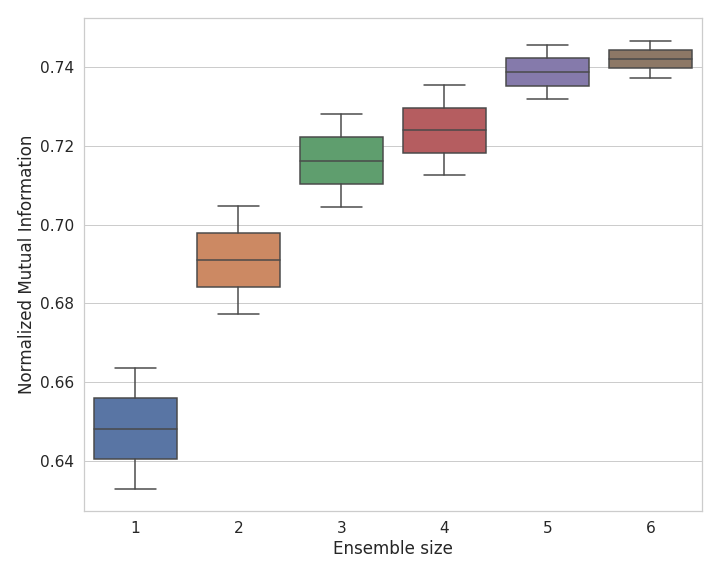}
  \vspace{-5mm}
  \caption{MNIST -- Impact of ensemble size}
  \label{fig:mnist_ensemble_size}
\end{figure}

\par
Figure~\ref{fig:mnist_ensemble_size} provides further empirical proof of the proposed method. As clearly indicated by conducted study -- including additional members in the ensemble steadily increases the models' performance. Results for the ensemble of size 5+ could also suggest an improvement in model stability. 

\par
Qualitative evaluation is conducted beyond classic evaluation using clustering metrics to understand the models' behaviour better. Figure~\ref{fig:cluster_inheritance} shows the latent space of the model, extracted just before the final clustering stage, using the TSNE~\cite{tsne} algorithm. For this purpose, a separate ensemble of 2 base members was created, latent spaces of which can be seen in Figures~\ref{fig:cluster_inheritance}(a) and~\ref{fig:cluster_inheritance}(b), respectively. Figure~\ref{fig:cluster_inheritance}(c) shows the latent space of the ensemble combining both models.

\begin{figure}[ht]
  \includegraphics[width=0.5\textwidth]{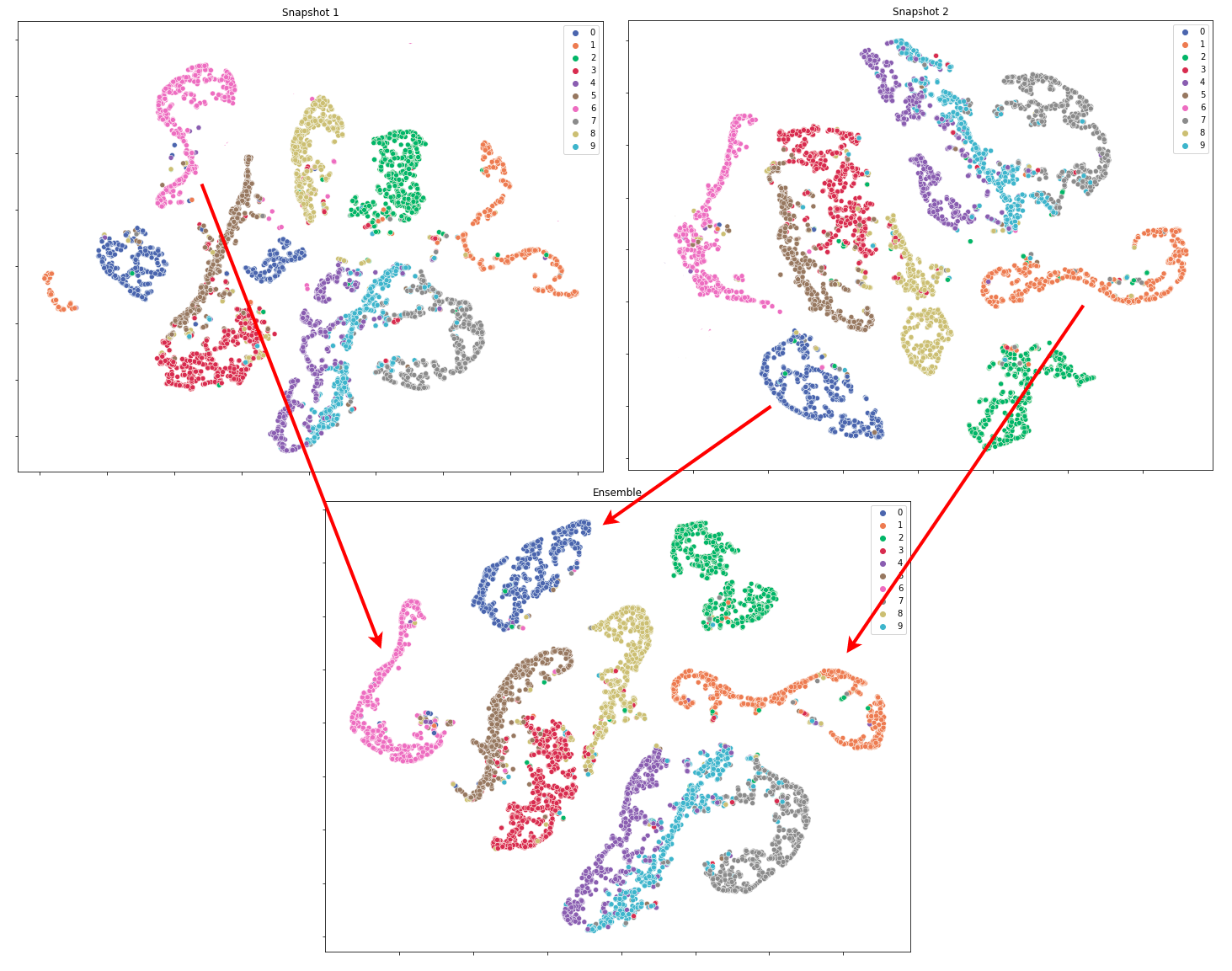}
  \caption{Ensemble cluster inheritance}
  \label{fig:cluster_inheritance}
\end{figure}

As seen in Figure~\ref{fig:cluster_inheritance}, an interesting example of ''cluster inheritance'' occurs. The latent space of the Snapshot 1 model has clearly deformed classes 0 and 1 -- they have irregular shapes, and most importantly, they are split into multiple pieces, interleaving other clusters. On the other hand, those classes are well formed in the latent space of the Snapshot 2 model, at the same time being clearly separated from other classes. Red arrows show cluster shapes of classes 0 and 1 generally preserved and transmitted into the final ensembles latent space. Similarly, the shape of class 6 is inherited from Snapshot 1 model.

\subsection{CIFAR-10}

\label{sec:cifar_full_results}

As with the MNIST dataset, all KMeans variants perform poorly compared to other methods. The algorithm completely fails to capture the complexity of the underlying data manifold. LSC-based models deliver results that are not much better. 

\begin{figure}[ht]
  \includegraphics[width=0.5\textwidth]{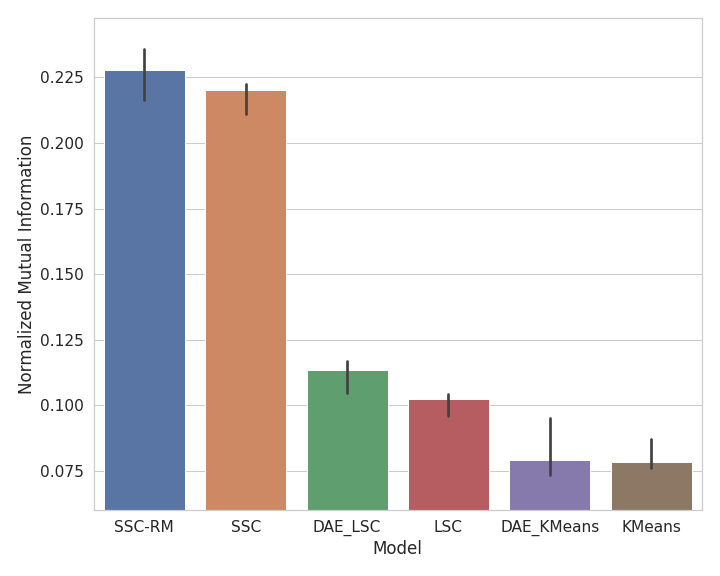}
  \vspace{-5mm}
  \caption{CIFAR-10 -- Models comparison}
  \label{fig:cifar_model_comparison}
\end{figure}

The advantage of the proposed algorithm is even more significant than observed on MNIST datasets -- here, the ensemble performs over twice as good as the base model. This behaviour supports the hypothesis that the method is better suited for more complex data that require bigger backbone networks. Above simply providing more descriptive encoding of data points, a network with more parameters responds better to a cosine learning scheduler. The model finds local minima of higher quality, thus providing greater diversity among base members and yielding better results. These experiments also confirm that a variant of the method with randomized metric (SSC-RM) can deliver even better results than the basic version of SSC.

\par
However, the low overall metric value observed in this dataset is worth mentioning. This can be explained by two main factors. First of all, the neural network used as the backbone is relatively small compared to the complexity of this problem. The second cause lies in the concept of unsupervised learning itself -- we do not provide a clear objective, what kind of similarities the algorithm should detect, so the model can ''choose to interpret'' the objective in a certain way. The quantitative evaluation using NMI disregards this factor, comparing model answers to a fixed set of classes. This is why qualitative research is essential to fully evaluate the model's performance.

\par
The inspection of created partitions and their representatives revealed interesting dependencies. Few clusters are correctly fitted to the original dataset's classes. However, as seen in Figure~\ref{fig:cifar_improper_classes}, clusters 1, 2 and 9 are not fitted to a single class, instead consisting of a mix of images. Yet, despite not being part of the same class, the images within a single cluster clearly manifest visual similarities.

\begin{figure}[ht]
  \includegraphics[width=0.5\textwidth]{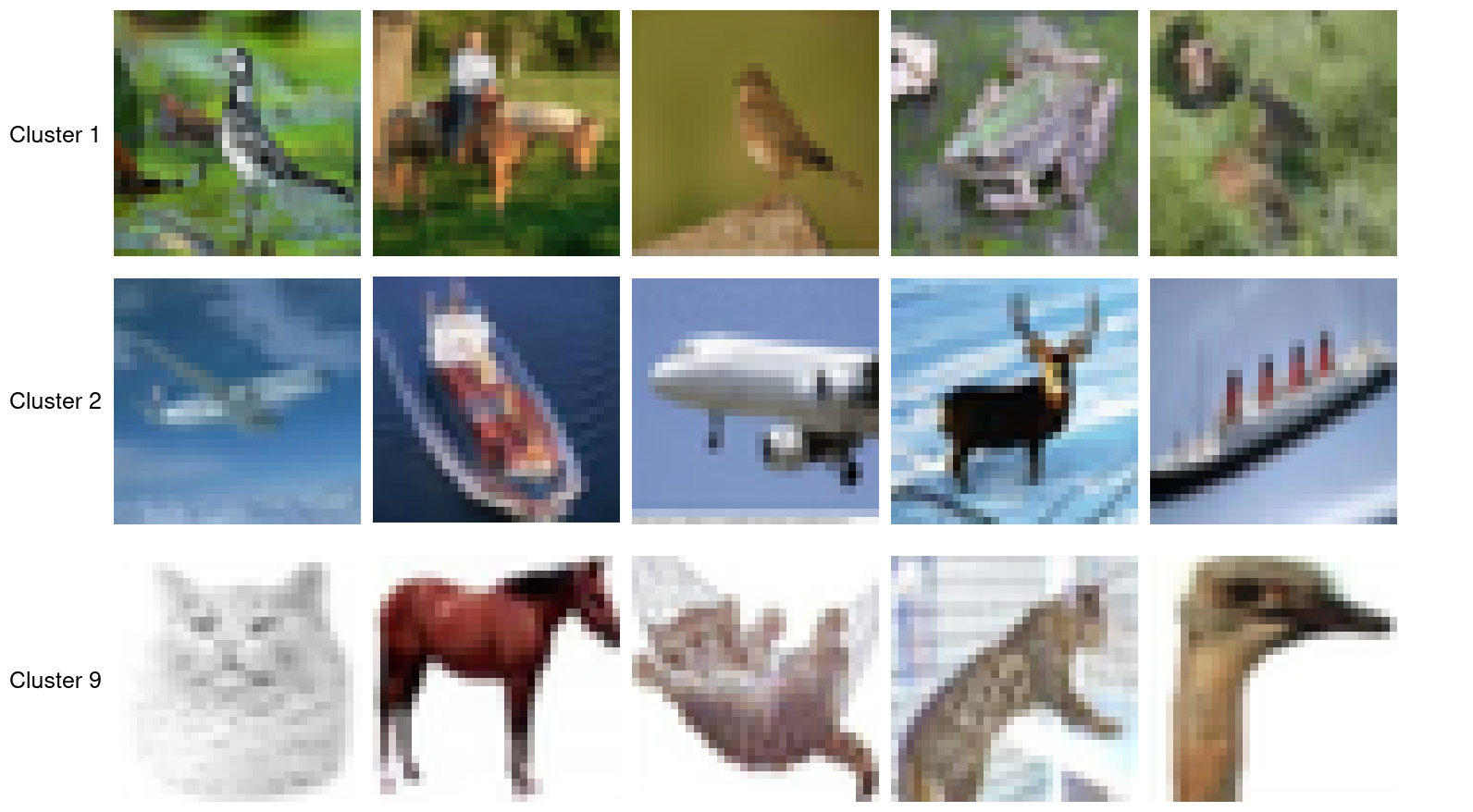}
  \caption{Final clustering -- misclassified images}
  \label{fig:cifar_improper_classes}
\end{figure}

Clusters seen in Figure~\ref{fig:cifar_improper_classes} are formed around dominating background colour -- green, blue and white in clusters 1, 2 and 9, respectively. This behaviour was to be expected, as the backbone network used in the experiment may not be expressive enough. Because of that, the extracted knowledge revolves around low-level features, such as background colour and not high-level semantic details. This explanation is further supported by the fact that this phenomenon did not occur while experimenting on the MNIST dataset -- there, the main object (digit) is presented without any background noise. The explanation of this behaviour also lies in the concept of unsupervised learning itself -- we do not provide a clear objective, what kind of similarities the algorithm should detect, so the model can ''choose to interpret'' the objective in a certain way. 

\section{Discussion and Conclusion}
\label{sec:conclusion}

In this work, a novel deep clustering ensemble algorithm was proposed -- Snapshot Spectral Clustering (SSC). It combines the descriptive power of Spectral Clustering with the robustness of ensemble methods, simultaneously reducing training costs by utilizing Snapshot Ensemble techniques. The method was designed to answer shortcomings discovered while reviewing of relevant literature, such as high resources requirements.

\par
Utilizing the Snapshot Ensemble methods~\cite{snapshot_ensemble} to create base members of the ensemble and generate diverse data encodings significantly reduced training costs, especially compared to the SC-EDAE~\cite{sc-edae} method. When in SC-EDAE, each model is trained independently, SSC allows to create \textit{m} base models with the same epoch budget as creating one model, thus reducing training cost \textit{m} times. Other minor improvements, such as using the MiniBatch KMeans~\cite{minibatch_kmeans} instead of the classic version to obtain landmarks, allowed further algorithm acceleration.

\par    
The use of the Snapshot Ensemble in an unsupervised learning paradigm is not yet found in literature, making it one of the main contributions of this paper. Also, the acceleration of the LSC landmark selection process with gradient-trained KMeans has not yet been proposed in other publications we know.

\par 
Transforming base embeddings using landmark-based representation~\cite{lsc} and utilising an efficient matrix fusion method leads to an outstanding reduction of memory requirements compared to the classic Spectral Clustering~\cite{spectral_clustering1} algorithm. Enforcing the sparsity of affinity matrix also allowed us to take advantage of special mechanisms designed for storing sparse matrices in memory, reducing those requirements even further\footnote{e.g. clustering 70,000 instances of the MNIST dataset, with Spectral Clustering requires the allocation of around 40GiB affinity matrix. SSC requires about 2MiB to process the same dataset.}.

\par
Additionally, a variant of the SSC method, called SSC-RM, was proposed. The random metric used in each base affinity matrix of the LSC algorithm allowed us to take advantage of different views of the data, enhancing the diversity among base members even further. This approach is not yet found in the literature, making it another innovative contribution of this paper.

\par
The conducted hyperparameter study provided valuable intuition to follow when selecting most of the values. It also pointed out more problematic hyperparameters while providing alternative approaches to tuning them. 

\par
\textbf{Future work}.
The proposed SSC algorithm delivers very promising results and exhibits potential for future work. As for setting the main direction for further improvement of the ensemble -- additional research is needed to choose a proper base model and backbone network architecture. Beyond obvious reasons, the sheer size of the underlying backbone network is an important factor that contributes to the SSCs' performance in more than one way:

\begin{enumerate}
    \item Bigger networks provide greater descriptive power, capable of discovering complex dependencies in the data.
    \item Studies have shown that unsupervised representation learning benefits even more from bigger models than its supervised counterpart~\cite {simclr}.
    \item Number of local minima grows exponentially with the number of parameters of neural network~\cite{kawaguchi}, making it easier for the SSCs learning scheduler to discover them.
    \item Increasing the depth and width of the neural network can further improve the quality of local minima~\cite{local_minima}, which can directly translate to the performance of SSCs ensemble members.
\end{enumerate}

\par
The improved model should also be subjected to more extensive experiments, using bigger datasets serving as benchmarks for algorithms of this type (e.g. CIFAR-100 or ImageNet). Further study should include a broad hyperparameter examination, emphasising model sensitivity and robustness, as well as the scalability of this solution. The comparison with state-of-the-art clustering algorithms will determine its full potential.
\bibliography{main}

\end{document}